\documentclass[preprint,review,3p, 12pt]{elsarticle}

\usepackage{algorithm}
\usepackage{algorithmic}

\usepackage{amsmath}
\usepackage{amssymb}
\usepackage{amsthm}
\usepackage{color}
\usepackage{dcolumn}

\biboptions{numbers,sort&compress}
\usepackage[figuresright]{rotating}
\usepackage{graphicx, color, epsfig, dsfont, amssymb, amsmath, amsthm, amsfonts, amsbsy, mathrsfs, mathtools, delarray}
\usepackage[colorlinks, linkcolor=green]{hyperref}

\usepackage{booktabs} 

\usepackage{graphicx}
\usepackage{subfigure}

\usepackage{float}  
\usepackage{lipsum}
\makeatletter
\newenvironment{breakablealgorithm}
{
	\begin{center}
		\refstepcounter{algorithm}
		\hrule height.8pt depth0pt \kern2pt
		\renewcommand{\caption}[2][\relax]{
			{\raggedright\textbf{\ALG@name~\thealgorithm} ##2\par}%
			\ifx\relax##1\relax 
			\addcontentsline{loa}{algorithm}{\protect\numberline{\thealgorithm}##2}%
			\else 
			\addcontentsline{loa}{algorithm}{\protect\numberline{\thealgorithm}##1}%
			\fi
			\kern2pt\hrule\kern2pt
		}
	}{
		\kern2pt\hrule\relax
	\end{center}
}
\makeatother

\journal{Neurocomputing}
\begin{document}

\begin{frontmatter}



\title{Multi-vehicle Platoon Overtaking Using NoisyNet Multi-Agent Deep Q-Learning Network}

\author[XTU]{Lv He} \ead{lv_he7@163.com}
\cortext[Corr_auth]{Corresponding author.}

\address[XTU]{School of Automation and Electronic Information, Xiangtan University, Xiangtan 411105, China}

\begin{abstract}
With the recent advancements in Vehicle-to-Vehicle communication technology, autonomous vehicles are able to connect and collaborate in platoon, minimizing accident risks, costs, and energy consumption. The significant benefits
of vehicle platooning have gained increasing attention from the automation and artificial intelligence areas. However,
few studies have focused on platoon with overtaking. To address this problem,
the NoisyNet multi-agent deep Q-learning algorithm is developed in this paper, which the NoisyNet is employed to improve the exploration of the environment. 
By considering the factors of overtake, speed, collision, time headway and following vehicles, 
a domain-tailored reward function is proposed to accomplish safe platoon overtaking with high speed. Finally, simulation results show that the proposed method achieves successfully overtake in various traffic density
situations.


\end{abstract}


\end{frontmatter}

\section{Introduction} \label{Sec_1}
In recent years, autonomous vehicles (AVs) and their technologies have received extensive attention worldwide. Autonomous driving has stronger perception and shorter reaction time compared to human driving. There is no human driver behavior such as fatigue driving, and it is safer for long-distance driving.  By 
using advanced Vehicle-to-Vehicle communication technologies, AVs 
are able to share information with each other and cooperate in dynamic driving tasks.
Through sharing information about the environment~\cite{xu2022cobevt,chen2022model}, locations, and actions, it will improve the driving safety~\cite{7593308}, reduce the traffic congestion~\cite{9541185}, and decrease the energy consumption~\cite{9408233}.  Multi-vehicle collaboration and overtaking are two important topics for AVs.

\begin{figure}
	\begin{minipage}[t]{1 \linewidth}
 		\centering	\includegraphics[width=1\textwidth]{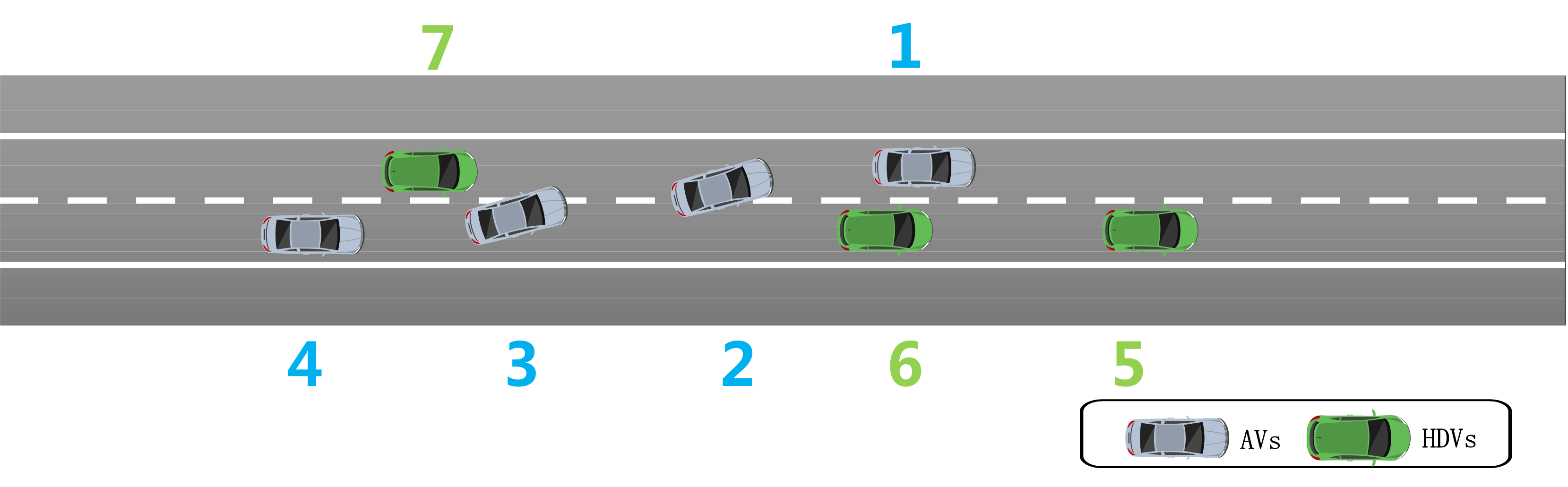}
	\caption{Illustration of the considered platoon overtaking traffic scenario. AVs (blue) and HDVs (green) coexist in the straight lane.}
		\label{fig1}
	\end{minipage}%
\end{figure}

\subsection{Multi-vehicle Collaboration} \label{Sec_1.1}

The cooperation of multiple vehicles is a promising way to improve traffic efficiency and reduce congestion, and many research works have been reported. By integrating CARLA~\cite{dosovitskiy2017carla} and SUMO~\cite{behrisch2011sumo},   OpenCDA~\cite{xu2021opencda,xu2023opencda} was proposed, which supports both cooperative driving automation prototyping and regular autonomous driving components. Based on OpenCDA, a series of multi-vehicle collaboration works focused on various fields were studied, e.g., collaborative perception~\cite{xu2022opv2v,10.1007/978-3-031-19842-7_7,xu2022cobevt,chen2022model,xu2022bridging,cai2022analyzing,li2022learning}, planning~\cite{han2022strategic}, localization, and safety system~\cite{xiang2022v2xp}. 
Coordinated strategies between autonomous vehicles could improve transportation efficiency and reduce unnecessary waiting time for passengers~\cite{9390363}. So that people have more time to do more meaningful things. Reasonable coordination of multiple vehicles in different road environments can improve traffic safety~\cite{9097378}, improve traffic efficiency~\cite{9497786}, and reduce fuel consumption~\cite{9575899}. These optimized indexes are of great significance to social traffic operations~\cite{9652467}. Based on the development of autonomous multi-vehicle coordination, in order to improve the transportation capacity of multiple vehicles, and reduce fuel consumption. A lot of useful techniques have been developed, and one of the very useful techniques is platoon~\cite{9103935}, which has been studied in great detail. Platoon driving refers to the situation where multiple vehicles coordinate and the rear vehicles follow the front vehicles at a short distance.

Reinforcement learning is a powerful method for decision-making, which has been applied to address autonomous driving problems in recent years. 
Multi-agent reinforcement learning algorithms can coordinate agents effectively and explore a large number of potential different environments quickly.  It not only enables multi-agent to adapt to the dense and complex dynamic driving environment but also enables multi-agents to make effective collaborative decisions~\cite{9700479}.
\subsection{Reinforcement Learning for Overtaking} \label{Sec_1.2}
Overtaking is an important way for AVs to improve driving efficiency, especially in mixed-traffic environments that contain AVs and human-driven vehicles (HDVs).  There are some works \cite{9304815, 8500718, 5710424} using reinforcement learning to handle single-vehicle overtaking problems. 
When making overtaking decisions, the agent needs to consider that other vehicles are in the vicinity of the agent and that different vehicles among them are traveling at different speeds. This requires agents to have multiple abilities to deal with overtaking problems~\cite{5710424}. Experienced human drivers can handle overtaking problems better, ~\cite{9561049} used curriculum reinforcement learning to make the agent perform overtaking operations comparable to experienced human drivers. When the platoon cannot travel at a relatively high or expected speed in the traffic flow, it will lead to a reduction in the efficiency of vehicle transportation. The platoon needs to speed up to overtake the slow vehicles in front and reach the destination faster. As shown in Fig.\ref{fig1}, platoon overtaking requires not only close coordination among members of the platoon but also the prevention of collision with other human vehicles around the platoon during the process of overtaking. The reinforcement learning algorithm with the more effective exploration of the environment can realize and improve the platoon's performance during overtaking.


However, to the best of our knowledge, using RL, especially multi-agent RL, for the AV platoon overtaking problem has rarely been studied. It is still an open and challenging problem, which motivates our studies in this paper.  
Inspired by NoisyNet~\cite{fortunato2018noisy}, we added the parameterized factorised Gaussian noise to the linear layer networks weights of the multi-agent deep Q-learning network, which induced stochasticity in the agent’s policy that can be used to aid efficient exploration. The factorised Gaussian noise parameters are learned by gradient descent along with the weights of the remaining networks. The longitudinal vehicle following distance in a platoon cannot be designed to be a fixed value. AVs at the end of a fixed-distance platoon may collide with nearby HDVs in overtaking passes, which will make the platoon much less safe when overtaking. Self-driving platoon faces the possibility of encountering other HDVs in the process of driving and overtaking, and need a suitable safety distance to adjust the self-driving vehicles policy to reduce the risk of collision. With the consideration of the above factors, a domain-tailored customized reward function is designed to achieve high-speed safe platoon overtaking. In order to reduce fluctuations arosen from the vehicle following reward, coefficients are added to the same lane following reward and the following distance interval reward, respectively. The total reward curve is more likely to converge after adding the coefficients, which is extremely helpful for multi-agents to learn stable policy. The contributions of the paper are summarized as follows:
\begin{itemize}
    \item The NoisyNet based multi-agent deep Q network (NoisyNet-MADQN) algorithm is developed for multi-vehicle platoon overtaking. By adding the parameterized factorised Gaussian noise to the linear layer networks weights of the multi-agent deep Q-learning network, which induced stochasticity in the agent’s policy that improves the exploration efficiency.  The  parameters of the factorised Gaussian noise are learned with gradient descent along with the remaining network weights. To reduce this computational overload, we select factorised Gaussian noise, which reduces the computational time for generating random numbers in the NoisyNet-MADQN algorithm. 
    
    \item By considering the factors of overtaking, speed, collision, time headway and vehicle following, the domain-tailored customized platoon overtaking is designed. The safety distance is designed in the vehicle-following reward to reduce the risk of collisions between the platoon and nearby HDVs while straight driving and overtaking. It is able to reduce collision rate of the AVs in the safety distance by the reward adaptively adjusting the distance they maintain from the AV in front. 
\end{itemize}

The rest of the paper is organized as follows. 
In Section II, the preliminary works of RL and the NoisyNet are presented. Section III presents  the reward function's design and the platoon overtaking algorithm. Section IV presents the experiments and results. Finally, Section V concludes our work.


\section{Problem Formulation} \label{Sec_2}
\subsection{Preliminary of Reinforcement Learning} \label{Sec_2.1}

In reinforcement learning, the agent's goal is to learn the optimal policy $\pi^*$ that maximizes the cumulative future rewards $R_t = \begin{matrix} \sum_{k=0}^T \gamma^k r_{t+k} \end{matrix}$, where $t$ is the time step, $r_{t+k}$ is the reward at time step $t+k$, and $  \gamma \in (0, 1] $ is the discount factor that quantifies the relative importance of future rewards. At time step $t$, 
the agent observes the state $s_t \in \boldsymbol{S} \subseteq \mathbb{R}^n$, selects an action $a_t \in \boldsymbol{A} \subseteq \mathbb{R}^m$, and receives a reward signal $r_t \subseteq \mathbb{R}$. $n$ represents the total number of agents, $i \in n$.

\subsubsection{Action Space}  \label{Sec_2.11}

An agent's action space $\boldsymbol{A_{i}}$ is defined as a set of high-level control decisions. Decision-making behaviors include  turning left, turning right, idling, speeding up, and slowing down. 


\subsubsection{State Space}  \label{Sec_2.12}

The state of agent ${i}$, $\boldsymbol{S}_{i}$, is defined as a
matrix of dimension $\boldsymbol{N_{\mathcal{N}}}_i \times \boldsymbol{m}$, where $\boldsymbol{N_{\mathcal{N}}}_i$ is the number of observed vehicles and $\boldsymbol{m}$ is the number of features. $\boldsymbol{Ispresent}$ is a binary variable that indicates whether there are other observable vehicles in the vicinity of 150 meters from the ego vehicle. $\boldsymbol{x}$ is the observed longitudinal position of the vehicle relative to the ego vehicle. $\boldsymbol{y}$ represents the lateral position of the observed vehicle relative to the ego vehicle. $\boldsymbol{v_{x}}$ and $\boldsymbol{v_{y}}$ represent the longitudinal and lateral speeds of the observed vehicle relative to the ego vehicle, respectively.

In the highway simulator, we assume that the ego vehicle can only obtain information about neighboring vehicles within 150 meters of the longitudinal distance of the ego vehicle. 
In the considered two-lane scenario (see Fig .\ref{fig1}), the neighboring vehicle is located in the lane and its neighboring lanes closest to the ego vehicle, with the $2^{th}$ AV as the ego vehicle and its neighboring vehicles as the $1^{th}$ AV, $3^{th}$ AV, $6^{th}$ HDV, and $7^{th}$ HDV.

\subsubsection{Reward Distribution}  \label{Sec_2.13}

In this paper, the NoisyNet multi-agent deep Q-learning is developed. As a multi-agent algorithm, since the vehicles in the platoon are the same type of vehicles, we assume that all the agents share the same network structure and parameters.  Our algorithm aims to maximize the overall  reward. To solve the communication overhead and credit assignment problems \cite{sutton2018reinforcement}, we use the following local reward design \cite{chen2021deep}.
So, the reward for the $i^{th}$ agent at time $t$ is defined as:
\begin{equation}\label{eq_3.14}
 r_{i,t} =  \frac{1} {\left|{V_i}\right| } \sum_{j \in V_i} r_{j,t},
\end{equation}
where $ \left| V_i \right|$ 
denotes the cardinality of a set containing the ego vehicle and its close neighbors.  This reward design includes only the rewards of the agents most relevant to the success or failure of the task \cite{elsayed2021safe}.

\subsection{Multi-agent Reinforcement Learning (MARL) and NoisyNets} \label{Sec_2.2}

This subsection will focus on the  NoisyNets and MARL. 
In the NoisyNet, its neural network weights and biases are perturbed by a function of noise parameters. These parameters are adjusted according to gradient descent ~\cite{fortunato2018noisy}. They assume that $y = f_\theta(x)$ is a neural network parameterized by a vector of noise parameters $\theta$ that accepts input $x$ and output $y$. In our experiments, we assume that there are $n$ AVs in the experimental environment. Each AV represents an agent, the $i^{th}$ AV represents the $i^{th}$ agent, $i \in n$.  $x_i$ represents the observed state of the $i^{th}$ agent, $y_i$ represents the action of the $i^{th}$ agent.
The noise parameter $\theta_i$ is denoted as $\theta_i \overset{def}{=} \mu_i + \Sigma_i \odot \varepsilon_i$,  $\zeta_i \overset{def}{=} (\mu_i , \Sigma_i)$ is a set of learnable parameter vectors, $\varepsilon_i$ is a zero-mean noise vector with fixed statistics, and $\odot$ denotes element multiplication. The loss of the neural network is wrapped by the expectation of the noise $\varepsilon_i: \bar{L} (\zeta_i) \overset{def}{=}  \mathbb{E} [L(\theta_i)]$. Then, the set of parameters $\zeta_i$ is optimized.
Consider the linear layers of the neural networks with $p$ inputs and $q$ outputs in these experiments, represented by
\begin{equation}\label{eq_2.1}
y_i = w_{i}x_{i} + b_{i},
\end{equation}
where $x_i \in \mathbb{R}^p$ are the layers inputs, $w_i \in \mathbb{R}^{q \times p}$ the weight matrix, and  $b_i \in \mathbb{R}^q$ the bias. The corresponding noisy linear layers are defined as:
\begin{equation}\label{eq_2.2}
y_i \overset{def}{=} (\mu_i^{w_i} + \sigma_i^{w_i} \odot \varepsilon_i^{w_i})x_{i} + \mu_i^{b_i} + \sigma_i^{b_i} \odot \varepsilon_i^{b_i},
\end{equation}
where $\mu_i^{w_i} + \sigma_i^{w_i} \odot \varepsilon_i^{w_i}$ and $\mu_i^{b_i} + \sigma_i^{b_i} \odot \varepsilon_i^{b_i}$ replace correspondingly $w_{i}$ and $b_{i}$ in Eq.(\ref{eq_2.1}). The parameters $\mu_i^{w_i} \in \mathbb{R}^{q \times p}$, $\mu_i^{b_i} \in \mathbb{R}^q$, $\sigma_i^{w_i} \in \mathbb{R}^{q \times p}$, $\sigma_i^{b_i} \in \mathbb{R}^q$, are learnable whereas $\varepsilon_i^{w_i} \in \mathbb{R}^{q \times p}$ and $\varepsilon_i^{b_i} \in \mathbb{R}^{q}$ are
noise random variables.
DeepMind introduced two types of Gaussian noise: independent Gaussian noise and factorised Gaussian noise. The computation overhead for generating random numbers in the algorithm is particularly prohibitive in the case of single-thread agents. To reduce the computation overhead for generating random numbers in the multi-agent
deep Q-learning network, we selected factorised Gaussian noise.

We factorize  $\varepsilon_{j,k}^{w}$, use $p$ unit Gaussian variables $\varepsilon_j$ for the noise of the inputs and $q$ unit Gaussian variables $\varepsilon_k$ for the noise of the outputs. 
Each $\varepsilon_{j,k}^{w}$ and $\varepsilon_{k}^{b}$ can then be written as:
\begin{equation}\label{eq_2.3}
\varepsilon_{j,k}^{w} = f(\varepsilon_j)f(\varepsilon_k),
\end{equation}
\begin{equation}\label{eq_2.4}
\varepsilon_{k}^{b} = f(\varepsilon_k),
\end{equation}
where $f$ is a real-valued function. In this experiment, we used
$f(x)=\text{sgn}(x) \sqrt{\left| x \right|}$.
We can obtain the loss of multiple noise networks. $\bar{L_i} (\zeta_i) = \mathbb{E}[L_i(\theta_i)]$, present the expectation of multiple gradients can be obtained directly from:

\begin{equation}\label{eq_2.5}
 \nabla\bar{L_i} (\zeta_i) =\nabla \mathbb{E}[L_i(\theta_i)] = \mathbb{E}[\nabla_{\mu_i,\Sigma_i}L(\mu_i + \Sigma_i \odot \varepsilon_i)].
\end{equation}
Using a Monte Carlo approximation to the above gradients, taking  samples $\xi_i$ at each step of optimization:

\begin{equation}\label{eq_2.6}
\nabla \bar{L_i} (\zeta_i) \approx \nabla_{\mu_i,\Sigma_i}L(\mu_i + \Sigma_i \odot \xi_i).
\end{equation}

In this work, we will no longer use $\epsilon$-greed, The policy greedily optimizes the (randomised) action-value function. Then the fully connected layers of the value network are parameterized to the noisy network, where the parameters are extracted from the noisy network parameter distribution after each replay step. 
Before each action, the noisy network parameters will be resampled, so that each action step of the algorithm can be optimized. In the target networks, the parameterized action-value function $Q(s_i, a_i,\varepsilon_i;\zeta_i)$ and $Q(s_i, a_i,\varepsilon^{'}_i;\zeta^{-}_i)$ can be regarded as a random variable when the linear layers in the network are replaced by the noisy layers. The outer expectation is with respect to the distribution of the noise variables $\varepsilon$ for the noisy value function $Q(s_i, a_i,\varepsilon_i;\zeta_i)$ and the noise variable $\varepsilon^{'}$ for the noisy target value function $Q(s_i, a_i,\varepsilon^{'}_i;\zeta^{-}_i)$.
So the NoisyNet-MADQN loss:

\begin{equation}\label{eq_2.7}
 \bar{L_i} (\zeta_i) =  \mathbb{E} \left[ \mathbb{E}_{(s,a,r,s_{t+1}) \sim D} [r + \gamma \mathbf{max} Q(s_i,a_i,\varepsilon^{'}_i;\zeta^{-}_i) - Q(s_i,a_i,\varepsilon_i;\zeta_i)]^2 \right].
\end{equation}

\section{NoisyNet-MADQN for platoon overtaking} \label{Sec_4}

\subsection{Reward function design} \label{Sec4_.1}

This subsection proposes a novel reward function for reinforcement learning algorithms to implement platoon overtaking. Reward functions are crucial for reinforcement learning models. By designing the reward function, we can guide the learning of RL agents to achieve our purpose.

\subsubsection{The overtake and speed evaluation} \label{Sec4_.2}

The vehicles will choose to drive at high speed driven by the reward, which will improve efficiency and allow more vehicles to reach their destination faster. When the speed of the front HDVs is less than the AVs, the platoon leader will increase the speed to overtake the front slower vehicles to get more rewards for completing the overtaking behavior. The other AVs in the platoon will also overtake the low-speed HDVs in front of them because of the following reward and speed reward. They follow the leader closely to form a platoon overtake. So the speed reward can also be seen as an overtaking reward. Therefore we still define the overtaking and speed reward for this vehicle as follows:

\begin{equation}\label{eq_4.5}
r_{os} =  \frac {v_t - v_{min}} {v_{max} - v_{min}} ,
\end{equation}
where ${v_{t}}$, ${v_{min}}$ = 20 m/s, and ${v_{max}}$ = 30 m/s are
the current, minimum, and maximum speeds of the
ego vehicle, respectively.

\subsubsection{ The collision penalty design }  \label{Sec_4.6}

Safety is the most critical factor in autonomous driving: if a collision occurs, the collision evaluation $\boldsymbol{r_{c}}$ is set to -1. If there is no collision, $\boldsymbol{r_{c}}$ is set to 0. The collision evaluation is defined as

 \begin{equation}\label{eq_4.6}
r_c =
\begin{cases}
-1,& \text{  collision;}\\
0,& \text{ safety}.
\end{cases}
\end{equation} 

\subsubsection{The time headway evaluation }  \label{Sec_4.7}
The time headway evaluation is defined as
\begin{equation}\label{eq_4.7}
 r_h = log \frac {d_{headway}} {t_h v_t},
\end{equation}
where $d_{headway}$ is the distance headway and $t_h$ is a
predefined time headway threshold. As such, the ego
vehicle will get penalized when the time headway
is less than  $t_h$ and rewarded only when the time
headway is greater than  $t_h$. In this paper, we choose
 $t_h$ as 1.2 s as suggested in \cite{ayres2001preferred}.

\subsubsection{The vehicles following evaluation }  \label{Sec_4.4}
 To keep the AVs in the platoon. So, the vehicles following evaluation is defined as 
 \begin{equation}\label{eq_4.8}
r_f =
\begin{cases}
0.3k_1{\left| C[i+1].p[0] - C[i].p[0] \right|  }/({t_h v_{max})}  ,& \text{    $\left| C[i+1].p[0] - C[i].p[0] \right|$ $\leq$  2$v_{max}$; }\\
0.7k_2,& \text{ $C[i].p[1] = C[i+1].p[1]$}  ,
\end{cases}
\end{equation} 
where $c[i]$ represents the $i^{th}$ AV, and $p[0]$,$p[1]$ represents the longitudinal and lateral coordinates of the AV, respectively.
\begin{figure}
	\begin{minipage}[t]{1 \linewidth}
 		\centering	\includegraphics[width=1\textwidth]{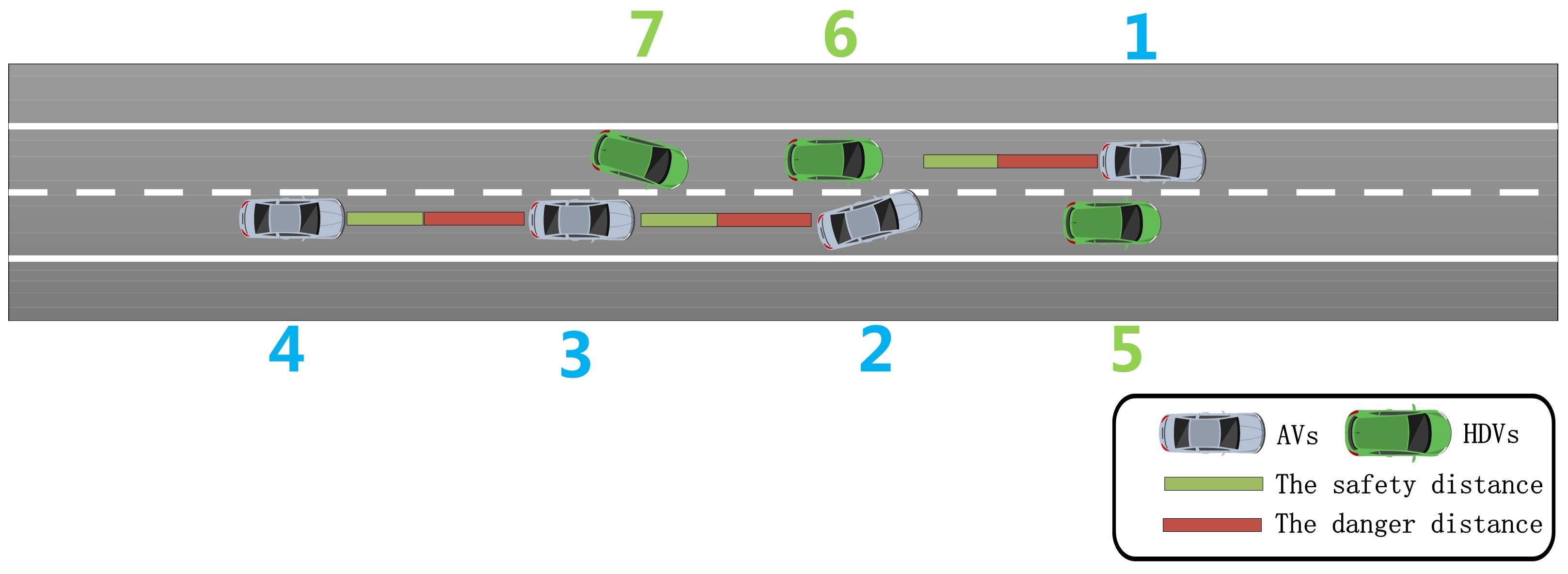}
	\caption{Illustration of the considered platoon overtaking traffic scenario. AVs (blue) and HDVs (green) coexist in the straight lane. The long green bar and the long red bar represent the safety distance and the danger distance respectively}
		\label{fig2}
	\end{minipage}%
\end{figure}
When the longitudinal distance between the AVs and the vehicle ahead of it in the platoon at time $t$ is kept within 60 $m$, the reward obtained at this time is $0.3k_1{\left| C[i+1].p[0] - C[i].p[0] \right|}/({t_h v_{max}})$. According to the time headway, we defined the danger distance as the distance range when the vehicle travels in a straight line at maximum speed for 1.2 $s$. 
The safety distance is defined as the distance range of 1.2$\sim$2 $s$ when the vehicle travels in a straight line at the maximum speed. As shown in Fig .\ref{fig2}.  
When the AVs are at the safety distance, the AVs get more rewards. When the distance between AVs is at 
 the danger distance, the closer the two AVs are to each other, the less reward they will receive. They also get the penalty because of the time headway evaluation.
The situations encountered by the self-driving platoon are classified into two types: platoon straight driving and platoon overtaking. HDVs can have an impact on the safety of the platoon while straight driving and overtaking. In Fig .\ref{fig2}, when the $7^{th}$ HDV enters between the $2^{th}$ AV and the $3^{th}$ AV, the fixed following distance of the rear AVs in order to follow the front AVs when the self-driving platoon leader accelerates will increase the risk of collision between the $3^{th}$ AV and the $7^{th}$ HDV. When the platoon leader in the self-driving platoon overtakes the $6^{th}$ HDV to obtain a greater overtake and speed reward. The AVs behind it follow closely. The $2^{th}$ AV and the $6^{th}$ HDV are very close to each other, and if the following distance is designed as a constant value, then they have a very high chance of collision. Based on the above, so we designed the safety distance to reduce the risk of following when overtaking. AVs can automatically adjust their following distance through training. After the training, the self-driving platoon vehicles will choose the safer and more effective action at time $t$  when faced with such situations. 

When the rear AVs in the platoon follow the front AVs in the same lane, at time $t$  the reward obtained is 0.7. Set the reward for keeping the AVs in the same lane to be larger than the reward for keeping the safety distance between the platoon. This is to ensure that when the platoon leader changes lanes to overtake, the following AVs can also change lanes in time to overtake. 

In our training, an episode has 100 steps, and the total reward accumulated by the acquisition fluctuates too much, which is not conducive to the agent obtaining a stable policy. So we add the weight $k_1$ to the reward obtained by keeping the distance with the vehicle ahead and the weight $k_2$ to the reward function of keeping in the same lane with the vehicle ahead in the platoon, respectively. $k_1$ and $k_2$ have values of 0.25 and 0.3, respectively. The fluctuation of the total reward becomes smaller after we add the weights, and it is easier for the agent to obtain a stable policy. Designing the weight $k_2$ to be slightly larger than $k_1$ can guide the vehicles behind the platoon leader to keep up with the platoon leader when the platoon leader overtakes. The smaller $k_1$ enables AVs to take advantage of the safety distance to avoid the risk of collision when AVs in a platoon encounter nearby HDVs while overtaking.

\subsubsection{ Total reward }  \label{Sec_4.5}

The reward function $\boldsymbol{r_{i}}$ is necessary for training multiple agents to behave as we desire. Since our goal is to keep the platoon while overtaking other vehicles. Therefore, the reward for the $\boldsymbol{i}^{th}$ agent at time step $\boldsymbol{t}$ is defined as follows:

\begin{equation}\label{eq_4.3}
r_{i,t} = w_{c}r_{c} + w_{os}r_{os} + w_{h}r_{h} + w_{f}r_{f}.
\end{equation}

Among them, $\boldsymbol{w_{c}}$, $\boldsymbol{w_{os}}$, $\boldsymbol{w_{h}}$, and $\boldsymbol{w_{f}}$ are the positive weight scalars corresponding to the collision assessment $\boldsymbol{w_{c}}$, the overtake and speed evaluation  $\boldsymbol{w_{os}}$,  the time headway evaluation $\boldsymbol{w_{h}}$, and AVs following evaluation $\boldsymbol{w_{f}}$, respectively. Since safety is the most important criterion, we made the $\boldsymbol{w_{c}}$ heavier than others. $\boldsymbol{w_{f}}$ is second only to $\boldsymbol{w_{c}}$ and higher than the other two weights. The coefficients $\boldsymbol{w_{c}}$, $\boldsymbol{w_{os}}$, $\boldsymbol{w_{h}}$, and $\boldsymbol{w_{f}}$ for the reward function are set as 200, 1, 4, and 5.

\subsection{ Noisy network Multi-Agent
Deep Q-learning Network }  \label{Sec_4.3}

\begin{figure}
	\begin{minipage}[H]{1 \linewidth}
 		\centering	\includegraphics[width=0.8\textwidth]{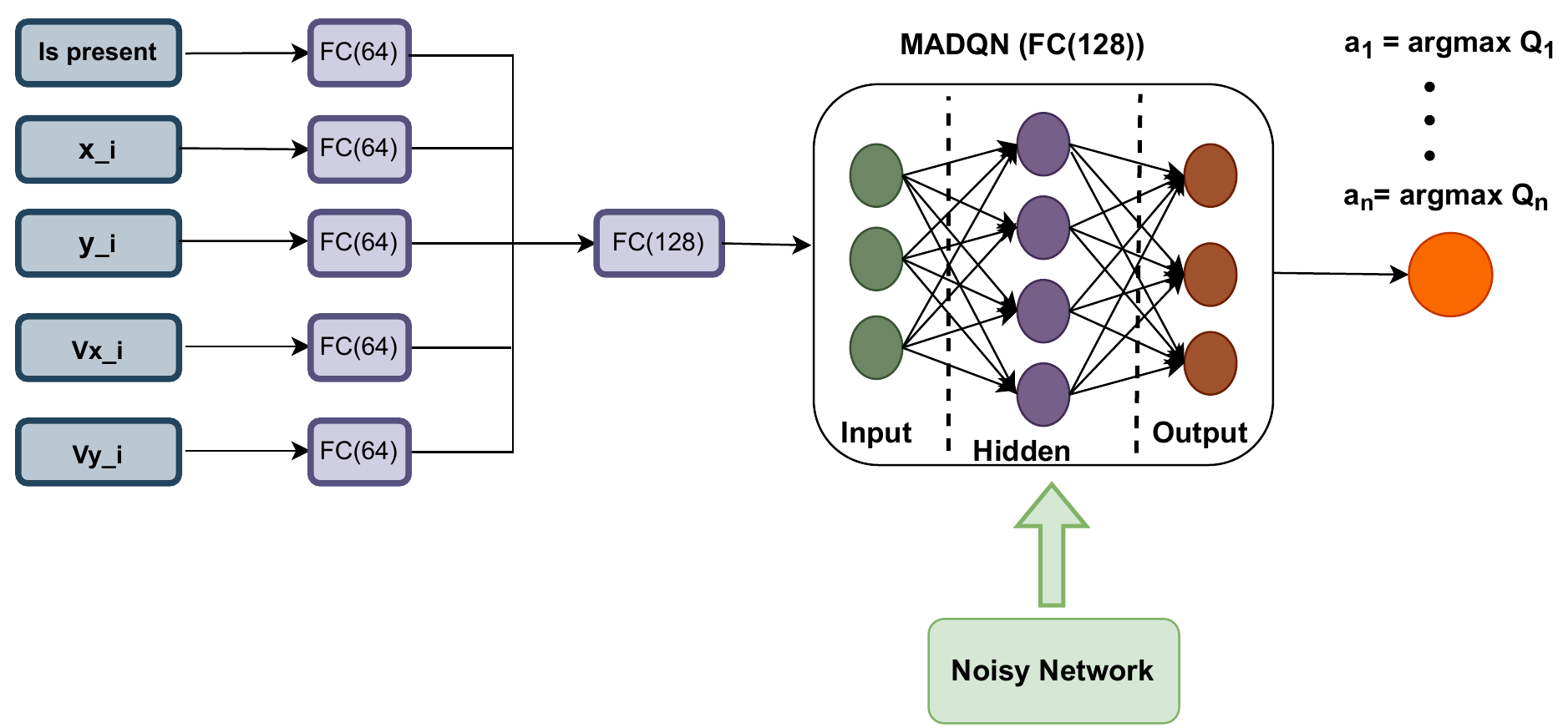}
	\caption{The  structure of the NoisyNet-MADQN network design is proposed, $i = 1, 2, ..., N$. And the numbers in parentheses indicate the size of the layers. }
		\label{fig3}
	\end{minipage}%
\end{figure}

 The leader of the platoon needs to choose the optimal policy $\pi^*$ when maintaining the platoon with the followers during overtaking, which improves the safety and efficiency of driving during overtaking. In the platoon overtaking scenario, we added the
parameterized factorised Gaussian noise to linear layer network weights of the multi-agent deep Q-learning network. Factorised Gaussian noise parameters are learned by gradient descent together with the remaining network weights. The computational overhead associated with single-thread agents is especially prohibitive. To overcome this computational overhead we select factorised Gaussian noise, which reduces the computational time for generating random numbers in the multi-agent deep Q-learning network. Fig.\ref{fig3} shows the network structure of our algorithm, in which states separated by physical units are first processed by separate 64-neuron fully connected (FC) layers. Then all hidden units are combined and fed into the 128-neuron FC layer. Based on this method, each agent in the multi-agent improves the exploration of the environment.  Our algorithm can obtain a larger optimal Q-value function for the same level of traffic compared to the original algorithm to achieve a better policy. More rewards for  the platoon also show that our algorithm can do better than the baseline in platoon overtaking. Algorithm \ref{alg:algorithm1} is the detailed procedure of NoisyNet-MADQN.

~\\
\begin{breakablealgorithm}
    \caption{NoisyNet-MADQN}
    \label{alg:algorithm1}
    \begin{algorithmic}[1] 
        \STATE Initialize replay buffers $D_1,...,D_n$ to the capacities $N_1,...N_n$,
action-value functions $Q_{1},...Q_{n}$ 
and target action-value functions $\hat{Q}^{(1)},...,\hat{Q}^{(n)}$ , $\varepsilon$ set of random variables of the network, $\zeta$ initial network parameters, $\zeta^-$ initial target network parameters, $N_T$ training batch size, $N^-$ target network replacement frequency
        \FOR {$episode$ =   \{1, ...$M$\} }
            \STATE  Initialize state $s_0^{(1),..,(n)}$ $\sim$ $Env$
            \FOR { $t$  = \{1, ...$T$\}}
                \STATE Set $s^{(1),..,(n)}$ $\gets$ $s_0^{(1),..,(n)}$
                \STATE Sample a noisy network $\xi_{1}$ $\sim$ $\varepsilon_{1}$, ...,$\xi_{n}$ $\sim$ $\varepsilon_{n}$
                \STATE Select $a_t^{1} = \mathbf{max}_a Q_{1}(s,a,\xi_{1};\zeta_{1}), ...,a_t^{(n)} = \mathbf{max}_a Q_{n}(s,a,\xi^{(n)};\zeta^{(n)})$   
                \STATE Execute joint action ($a_t^{(1)}$, ...,$a_t^{(n)}$ )  and
    observe reward $r^{(1)}_t, ...,r^{(n)}_t$ , and new state ($s^{(1)}_{t+1}, ..., s^{(n)}_{t+1}$)
        
                \STATE Store transition($s_t^{(1)}$, $a_t^{(1)}$, $r_t^{(1)}$,$s_{t+1}^{(1)}$, ..., $s_t^{(n)}$, $a_t^{(n)}$, $r_t^{(n)}$,$s_{t+1}^{(n)}$) in $D_1, ..., D_n$
                \STATE Sample a minibatch of $N_T$ transitions (($s_j^{(1)}$,$a_j^{(1)}$,$r_j^{(1)}$,$s_{j+1}^{(1)}$) $\sim$ $D_1)_{j=1}^{N_T}$, ...,(($s_j^{(n)}$,$a_j^{(n)}$,$r_j^{(n)}$,$s_{j+1}^{(n)}$) $\sim$ $D_n)_{j=1}^{N_T}$
                \STATE Sample the noisy variable for the online network $\xi_{1}$ $\sim$ $\varepsilon_{1}$, ...,$\xi_{n}$ $\sim$ $\varepsilon_{n}$
                \STATE Sample the noisy variable for the target network $\xi^{'}_1$ $\sim$ $\varepsilon_{1}$, ...,$\xi^{'}_n$ $\sim$ $\varepsilon_{n}$
                \FOR{$j$ $\in$ \{1, ...,$N_T$ \}}
                    \IF{$s_j$ is a terminal state}
                        \STATE   $\hat{Q}$ $\gets$ $r_j$
                    \ELSE 
                        \STATE $\hat{Q}$ $\gets$ $r_j$ + $\gamma$ $\mathbf{max}$ $Q$($s_j$,$a_j$,$\xi^{'}$ ; $\zeta^-$)
                    \ENDIF
                    \STATE Perform a gradient descent step on ($\hat{Q}_{1}$ - $Q_{1}(s_j^{(1)},a_j^{(1)}$,$\xi_{1}$;$\zeta_{1}))^2$, ..., ($\hat{Q}_{n}$ - $Q_{n}(s_j^{(n)},a_j^{(n)}$,$\xi_{n}$;$\zeta_{n}))^2$  
                \ENDFOR
                \IF{$t$ $\equiv$ $0$ $(mode N^-)$}
                    \STATE Update the target network $\zeta_{1}^- = \zeta_1$ , ..., $\zeta_{n}^- = \zeta_n$
                \ENDIF
            \ENDFOR
        \ENDFOR
    \end{algorithmic} 
\end{breakablealgorithm}

\section{Experiments and Discussion} \label{Sec_4}

In this section, the effectiveness of the proposed method is verified by simulation. Some implementation details of the experiments are given, and we evaluate the performance of the proposed MARL algorithm in terms of training effectiveness for overtaking in the considered road scenario shown in Fig .\ref{fig1}. The experimental results are also discussed.

For cost and feasibility considerations, we conducted experiments in the simulator.  We use an open-source simulator developed on highway-env~\cite{highway-env} and modify it as needed. The simulator is capable of simulating the driving environment and vehicle sensors. These vehicles randomly appear on the highway with different initial speeds of 20-30 m/s. And take random actions.

\subsection{Experimental Settings} \label{Sec4_4.1}

In order to fully demonstrate the effectiveness of our proposed method. Three traffic density levels were used to evaluate the effectiveness of the proposed method, corresponding to low, middle, and high levels of traffic congestion. We train the POMARL algorithm for 200 episodes by applying two different random seeds. The same random seed is shared among agents. These experiments were performed on an ubuntu server with a 2.7GHz Intel Core i5 processor and 16GB of RAM. The number of vehicles in different traffic modes is shown in Table \ref{table: 1}.

\begin{table}[h]\tiny\centering
	\begin{minipage}{0.7\linewidth}
		\centering
		\caption{\textbf{ Traffic density modes.}}
		\label{table: 1}
		\resizebox{1\textwidth}{!}{
			\begin{tabular}{lccc}
				\hline
				 Density  & AVs & HDVs & Explanation \\ \hline 
				  1 	& 4 & 1-2 & low level \\ 
				  2     & 4 & 2-3 &  middle level \\
			  	  3 	& 4 & 3-4 & high level \\ \hline
			\end{tabular}
		}
	\end{minipage}
	\hfill
\end{table}

\begin{figure*}[h]
\centering
\subfigure[Low level]
{
    \begin{minipage}[b]{0.31\linewidth}
        \centering
        \includegraphics[scale=0.4]{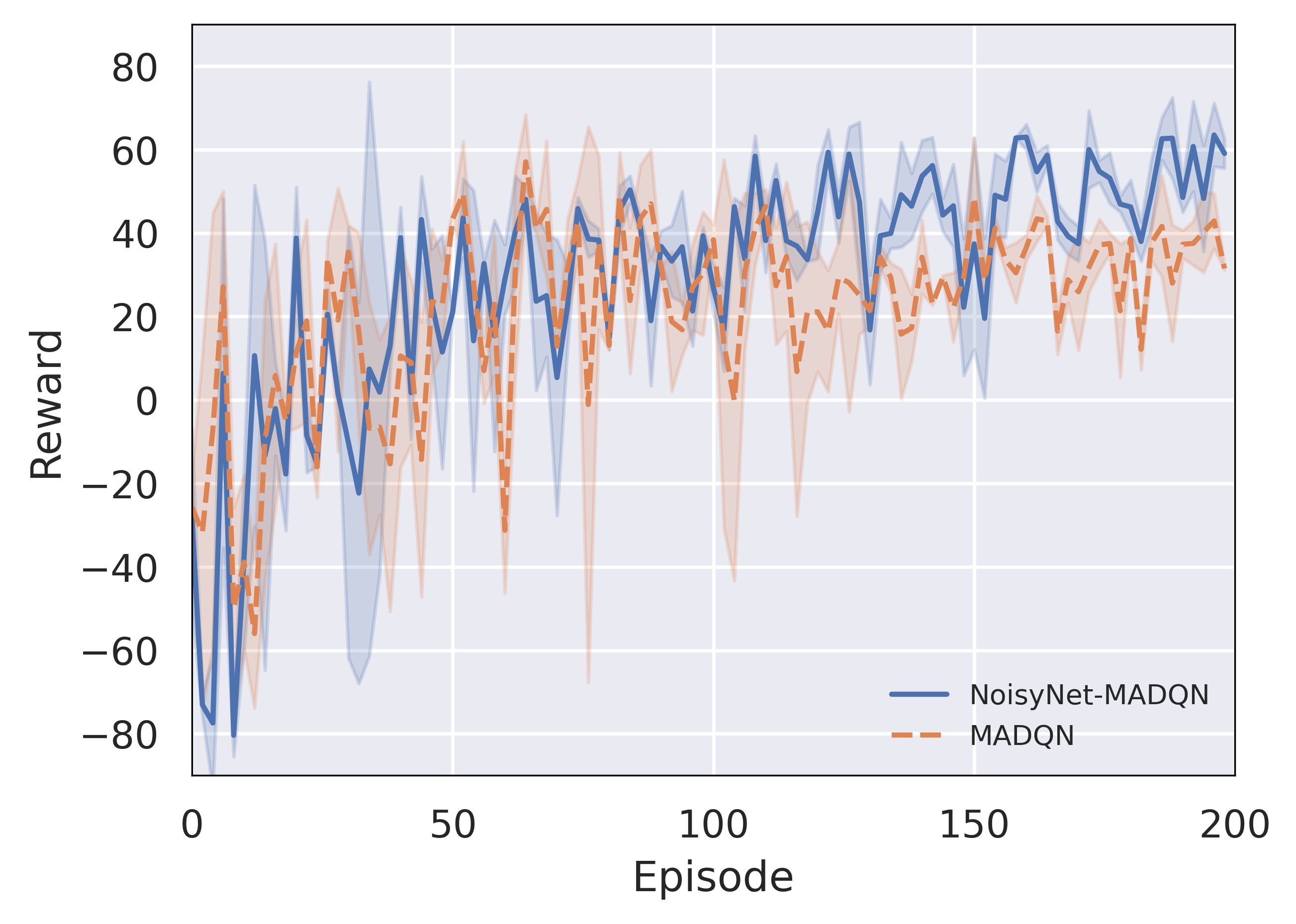}
    \end{minipage}
}
\subfigure[Middle level]
{
 	\begin{minipage}[b]{0.31\linewidth}
        \centering
        \includegraphics[scale=0.4]{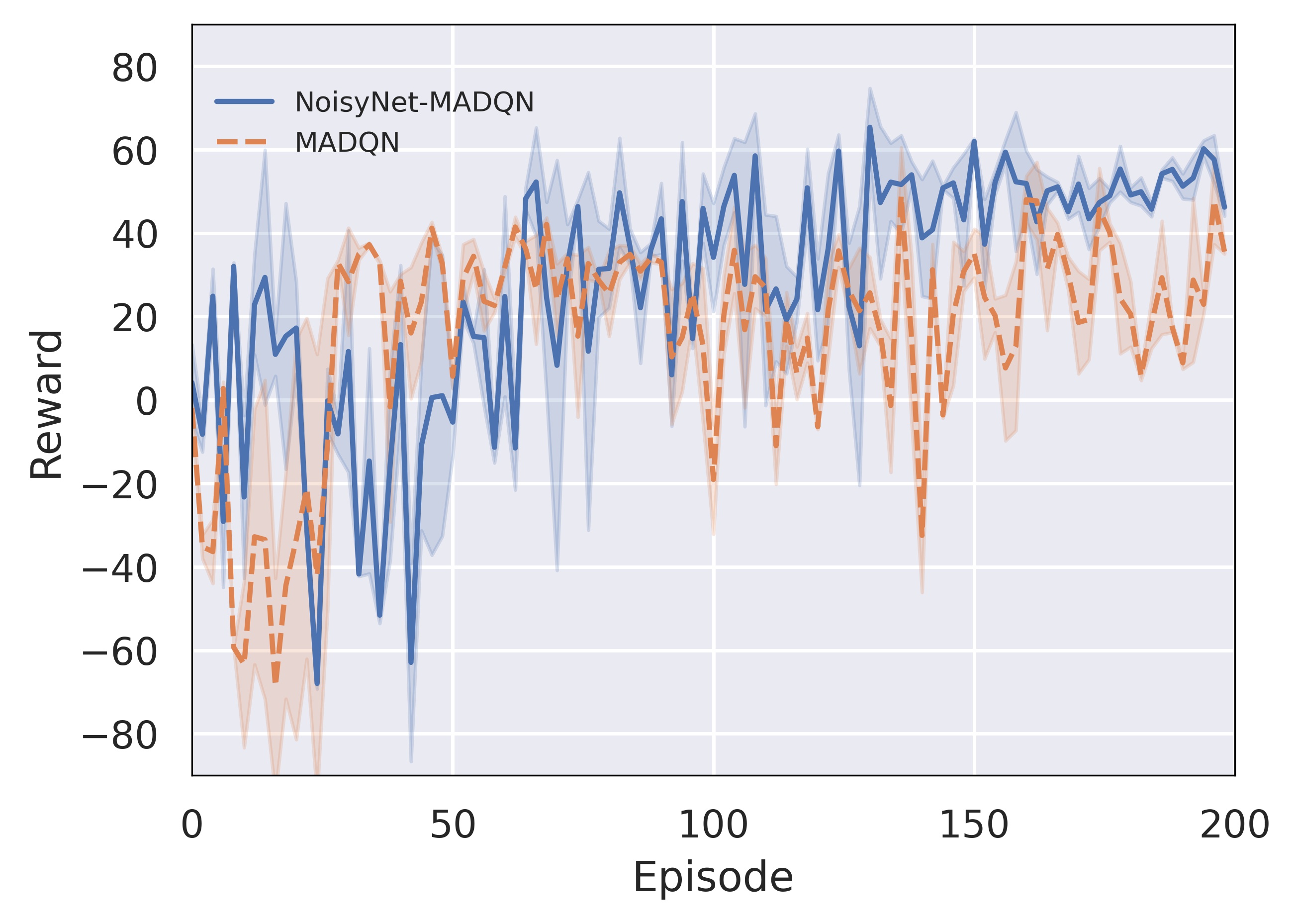}
    \end{minipage}
}
\subfigure[High level]
{
 	\begin{minipage}[b]{0.31\linewidth}
        \centering
        \includegraphics[scale=0.4]{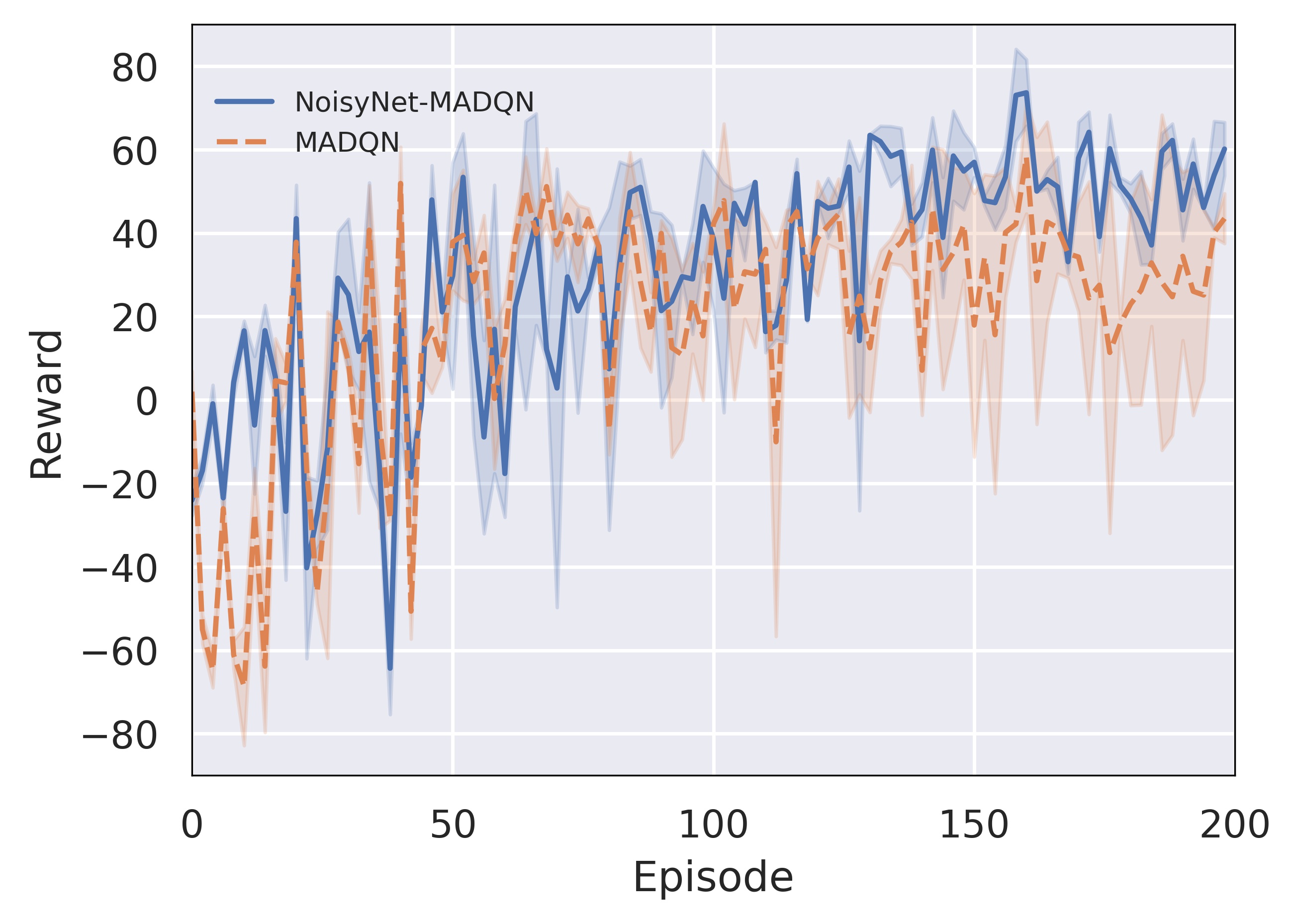}
    \end{minipage}
}
\caption{Evaluation curves during training with different algorithms for different traffic levels.}
\label{fig4}
\end{figure*}

\begin{figure}[!h]
\centering
\subfigure[Platoon overtaking at low level]{
\includegraphics[width=1\textwidth]{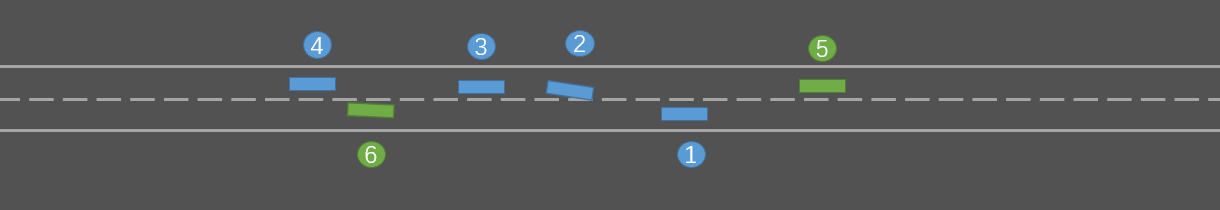}
}
\subfigure[Platoon overtaking at middle level]{
\includegraphics[width=1\textwidth]{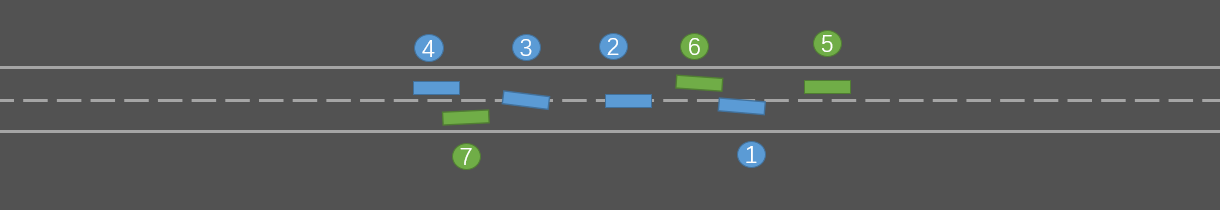}
}
\quad
\subfigure[Platoon overtaking at high level]{
\includegraphics[width=1\textwidth]{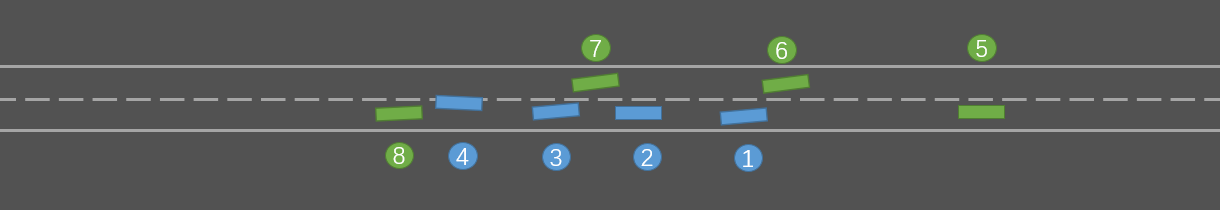}
}
\caption{Platoon overtaking in the simulation environment}
\label{fig5}
\end{figure}

 Fig .\ref{fig4} shows the comparison between our algorithm and the baseline algorithm. It shows that NoisyNet-MADQN performs better than MADQN. We put the rewards obtained by the two algorithms respectively under low-level, mid-level, and high-level to make a comprehensive curve comparison. We tabulate these reward values and divide the rewards within 200 episodes into five intervals. The value of each interval is their average. In Table \ref{table:2}, the NoisyNet-MADQN obtains higher rewards than MADQN most of the time.
 Fig .\ref{fig5} shows snapshots of the platoon overtaking at low, middle, and high levels, respectively. It shows that our method can be applied to solve platoon overtaking. From the comparison of NoisyNet-MADQN and baseline algorithm rewards, it can be found that the proposed new algorithm obtains better results in platoon overtaking. It shows that the platoon has achieved better results in both driving efficiency, safety, and convoy coordination.

\begin{table*}[!h]
\caption{Reward compare} \label{table:2}
\begin{tabular}{ll c c c c c}
\toprule
Density & Method & 1 $\sim$ 40 & 41 $\sim$ 80 & 81 $\sim$ 120 & 121 $\sim$ 160 & 161 $\sim$ 200 \\
\midrule
low level & NoisyNet-MADQN & -14.31 & \textbf{28.11} & \textbf{36.03} & \textbf{44.32} & \textbf{52.51} \\
          & MADQN          & \textbf{-5.02} & 23.47  & 28.02 & 27.95 & 33.48  \\
\addlinespace
middle level & NoisyNet-MADQN & \textbf{-6.13} & 12.33 & \textbf{34.38} &  \textbf{45.61} & \textbf{50.53}  \\
             & MADQN          & -15.75 & \textbf{28.51} & 19.77  & 17.48 & 29.01   \\
\addlinespace
high level & NoisyNet-MADQN & \textbf{-4.48} & 19.28 & \textbf{33.52}  & \textbf{51.58} & \textbf{52.34} \\
           & MADQN          & -19.13  & \textbf{28.62} & 26.60   &31.72 & 31.29	\\
\bottomrule
\end{tabular}
\end{table*}


\section{Conclusions} \label{Sec_6}

This paper proposes the NoisyNet multi-agent deep Q-learning network. The computation overhead for generating random numbers in the algorithm is particularly prohibitive in the case of single-thread agents. To reduce the computational overhead for generating random numbers in the multi-agent deep Q-learning network we selected factorised Gaussian noise. By adding the parameterized factorised Gaussian noise to the linear layer networks weights of the multi-agent deep Q-learning network, the induced randomness of the agent's policy can be used to help effective exploration. The parameters of the factorised Gaussian noise are learned with gradient descent along with the remaining network weights. To prove the effectiveness of our proposed algorithm, the proposed algorithm is compared with the baseline in the platoon overtaking tasks. By considering overtake, speed,
collision, time headway and following vehicles factors, a domain-tailored reward function is proposed to accomplish safe platoon overtaking with high speed. The safety distance in the vehicle following evaluation allows the vehicles in the platoon to adjust the following distance to avoid collision when the platoon faces the nearby HDVs inserting into the platoon. It also allows the rear vehicle in the platoon to avoid collision with the nearby HDVs when following the overtaking vehicle. We compared it with the existing baseline algorithm at three different traffic densities and showed that it performs better than the baseline, achieving reasonable results.




~\linebreak ~\linebreak
\noindent \textbf{}
\bibliographystyle{elsarticle-num}
\bibliography{References}






\end{document}